\documentclass{article}
\usepackage[english]{babel}
\usepackage[letterpaper,top=2cm,bottom=2cm,left=3cm,right=3cm,marginparwidth=1.75cm]{geometry}
\usepackage[utf8]{inputenc}
\usepackage{lmodern}
\usepackage[colorlinks=true, allcolors=black]{hyperref}
\usepackage[affil-it]{authblk}
\usepackage{graphicx}
\usepackage[acronym]{glossaries}
\usepackage[
backend=biber,
style=nature,
sorting=none
]{biblatex}
\usepackage{csquotes}
\usepackage{multirow}
\usepackage{multicol}

\title{pyAKI - An Open Source Solution to Automated KDIGO classification}
\author [1,*]{Christian Porschen}
\author[2,3,4*]{Jan Ernsting}
\author[5]{Paul Brauckmann}
\author[1]{Raphael Weiss}
\author[1]{Till Würdemann}
\author[1]{Hendrik Booke}
\author[1]{Wida Amini}
\author[1]{Ludwig Maidowski}
\author[2,4]{Benjamin Risse}
\author[3]{Tim Hahn}
\author[1]{Thilo von Groote}

\affil[1]{Department of Anaesthesiology, Intensive Care and Pain Medicine, University Hospital Münster}
\affil[2]{Institute for Geoinformatics, University of Münster, Germany}
\affil[3]{University of Münster, Institute for Translational Psychiatry, Germany}
\affil[4]{Faculty of Mathematics and Computer Science, University of Münster, Germany}
\affil[5]{FH Münster - University of Applied Sciences}
\affil[*]{These authors contributed equally to this work}

\makeglossaries
\newacronym{aki}{AKI}{Acute kidney injury}
\newacronym{uo}{UO}{urine output}
\newacronym{scr}{SCr}{serum creatinine}
\newacronym{gfr}{GFR}{glomerular filtration rate}
\newacronym{icu}{ICU}{intensive care unit}
\newacronym{amds}{AmsterdamUMCdb}{Amsterdam University Medical Center Database}
\newacronym{mimic}{MIMIC}{Medical Information Mart for Intensive Care}
\newacronym{kdigo}{KDIGO}{Kidney Disease Improving Global Outcomes}
\newacronym{mg}{mg}{milligram}
\newacronym{mmol}{mmol}{millimol}
\newacronym{ml}{ml}{milliliter}
\newacronym{kg}{kg}{kilograms}
\newacronym{dl}{dl}{deciliter}
\newacronym{mgdl}{mg/dl}{milligrams per deciliter}
\newacronym{mlmin}{ml/min}{milliliters per minute}
\newacronym{mlkgh}{ml/kg/h}{milliliters per kilogram per hour}
\newacronym{h}{h}{hour}
\newacronym{sql}{SQL}{Structured Query Language}
\newacronym{csv}{CSV}{Comma-separated value}

\begin{document}
\maketitle

\begin{abstract}
\acrfull{aki} is a frequent complication in critically ill patients, affecting up to 50\% of patients in the intensive care units. The lack of standardized and open-source tools for applying the \acrfull{kdigo} criteria to time series data has a negative impact on workload and study quality. This project introduces pyAKI, an open-source pipeline addressing this gap by providing a comprehensive solution for consistent \acrshort{kdigo} criteria implementation.

The pyAKI pipeline was developed and validated using a subset of the \acrfull{mimic}-IV database, a commonly used database in critical care research. We defined a standardized data model in order to ensure reproducibility. Validation against expert annotations demonstrated pyAKI's robust performance in implementing \acrshort{kdigo} criteria. Comparative analysis revealed its ability to surpass the quality of human labels.

This work introduces pyAKI as an open-source solution for implementing the \acrshort{kdigo} criteria for AKI diagnosis using time series data with high accuracy and performance. 
\end{abstract}

\section{Introduction}\label{sec:introduction}
\acrfull{aki} is a common organ dysfunction in critically ill patients, affecting up to 50\% of all \acrfull{icu} patients~\cite{hoste_epidemiology_2015, zarbock_epidemiology_2023}. Definitions for \acrshort{aki} changed over the years~\cite{chawla_acute_2017, bellomo_acute_2004}. Recently, the definition of \acrshort{aki} has been made easier, by implementing the \acrshort{kdigo}-definition for \acrshort{aki} ~\cite{levey_defining_2022}. In 2012, the \acrshort{kdigo}-Initiative provided a new standardized definition by defining \acrshort{aki} based on routinely available functional kidney markers using the \acrfull{uo} and \acrfull{scr} of patients~\cite{khwaja_kdigo_2012}. While this definition made it possible to investigate the epidemiology of this syndrome and the sequalae, there is currently no open software tool available to identify \acrshort{aki} events based on the \acrshort{kdigo} criteria in time series data. Annotating this data by trained physicians is time consuming and resource intensive. However, the development of clinical decision support systems requires large amounts of high quality labels, especially when using machine learning methods which typically require large amounts of data. Here, we provide an approach for standardizing and leveraging clinical data and diagnoses into software to facilitate and enable development of such decision support systems.

In recent years, databases were developed containing valuable patient data for development of clinical decision support systems~\cite{johnson_mimic-iv_2023, thoral_sharing_2021, hyland_early_2020, pollard_eicu_2018, sicdb_doc_main_2022}. However, using those databases for training of machine or deep learning based systems remains difficult due to missing annotation. Generating these annotations is highly time consuming and costly, as medical diagnoses have to be assigned by medical experts. Additionally, most databases use custom data formats, effectively hindering annotation. Developing systems for automatic assignment of diagnosis based on medical standards and verified by medical experts is a valuable next step in the direction of building medical decision support systems.

First efforts to develop automated annotation systems started recently with OpenSep. OpenSep, which has been developed in 2022, is one of the pipelines implementing Sepsis-3 guidelines into an open-source software, leveraging reusable and standardized concepts and implementing them in Python, a commonly used programming language~\cite{hofford_opensep_2022}. Thereby, users are enabled to use the implemented rules developed by medical domain experts rendering the subsequent results more reliable and generating a common definition of diagnostic criteria. Additionally, Python enables data scientists to implement those algorithms as "diagnostic feature extractors" into advanced data processing pipelines. 

So far, there has not been a single open source, ready to use pipeline software implementation, processing a standardized, simple data format into a standardized definition of \acrshort{aki}. While projects like the \acrshort{mimic}-database~\cite{johnson_mimic-iv_2023} or the \acrshort{amds}~\cite{thoral_sharing_2021} have recently undergone efforts to implement \acrshort{kdigo} criteria within their own data sets, a pipeline implementation for cross-project application defining a common standard for \acrshort{aki} is lacking. Furthermore, the validity of these annotations in comparison to annotations labelled by medical experts was not assessed.

Therefore, we propose pyAKI – an open-source, standardized, generalizable and clinically tested pipeline for implementing \acrshort{kdigo} criteria in time series data. Furthermore, we defined a data model to standardize the input required for the implementation of these guidelines (\autoref{fig:pyAKI_ERD}). This minimal data model is the required minimal subset of variables to produce valid \acrshort{kdigo} criteria based \acrshort{aki} diagnosis. The necessary data fields are present in the databases currently utilized in the field, enabling and facilitating access to valid diagnosis annotation. Additionally, we provide methods to support users in transforming their own raw data into a format usable by the pipeline. To evaluate pyAKI's performance, we created an expert annotation for a dataset extracted from the \acrshort{mimic}-IV database, reviewed by domain experts, that evaluated all cases at each subsequent hourly point in time, during the \acrshort{icu} stay. All labels are based on the \acrshort{kdigo} 2012 definition and were verified by an expert panel of physicians experienced in the care of patients with \acrshort{aki}.

The entire software we developed is open source and available for download ~\footnote{github.com/aidh-ms/pyAKI}. The dataset generated for validation and the software implementation are licensed under an open source license and publicly available ~\footnote{DOI: 10.17879/17988545762}.

\section{Materials and Methods}\label{sec:materials_methods}
\subsection{\acrshort{aki} Definition}\label{sec:aki_definition}
\acrshort{aki} was defined applying the \acrshort{kdigo} criteria using \acrshort{uo} and \acrshort{scr}~\cite{khwaja_kdigo_2012}. According to \acrshort{kdigo}, requirement of dialysis is defined as \acrshort{aki} stage 3. To facilitate this \acrshort{aki} criterion, input of dialysis data was implemented accordingly. From a computational perspective, the \acrshort{kdigo} criteria classify \acrshort{aki} via three different pathways: First, a reduction in \acrshort{uo}; second, an elevation of the kidney function marker \acrshort{scr} against a predefined baseline, which in turn can be a relative or an absolute increase; third the requirement of dialysis. According to the \acrshort{kdigo} criteria, a reduction in \acrshort{uo} below 0.5\acrshort{mlkgh} for 6 to 12 hours was defined as \acrshort{aki} stage 1. Reduction in \acrshort{uo} below 0.5\acrshort{mlkgh} for more than 12 hours was defined as \acrshort{aki} stage 2. Reduction of \acrshort{uo} below 0.3\acrshort{mlkgh} for at least 24 hours or anuria for at least 12 hours was defined as \acrshort{aki} stage 3.  \acrshort{scr} elevations have to be compared to a baseline in most cases. Our methods of defining and implementing baselines for \acrshort{scr} are explained below. A \acrshort{scr} rise relative to the baseline creatinine was classified as relative creatinine stage. An 1.5-1.9 fold increase relative to baseline \acrshort{scr} was defined as \acrshort{aki} stage 1, a 2-2.9 fold increase was defined as \acrshort{aki} stage 2, and a 3 fold increase was defined as \acrshort{aki} stage 3. An absolute \acrshort{scr} increase of 0.3\acrshort{mgdl} over baseline was classified as \acrshort{aki} stage 1. An absolute \acrshort{scr} increase over 4\acrshort{mgdl} irrespective of the baseline was classified as \acrshort{aki} stage 3. dialysis was also considered for classifying \acrshort{aki}. Any use of dialysis was classified as \acrshort{kdigo} stage 3. Definition of anuria varies in the literature. Therefore, we provide users the possibility to provide their own threshold for anuria based on body weight. An overview of the \acrshort{aki} definition according to the \acrshort{kdigo} criteria is depicted in \ref{tab:aki_staging}. In order to use the pipeline, users have to provide values for the weight of each patient. Users can decide for themselves whether an ideal body weight or adjusted body weight should be used.

\begin{table}
    \begin{tabular}{ |p{0.75cm}||p{5cm}|p{5cm}| p{2cm} | }
    \hline
    \acrshort{aki} Stage & Serum Creatinine & Urine Output & Dialysis \\
    \hline \hline
    1 & 1.5 - 1.9 fold increase or absolute elevation of $\geq$ 0.3 \acrshort{mgdl} relative to baseline& $\textless$ 0.5 \acrshort{mlkgh} for 6-12 hours & \\
    \hline
    2 & 2 - 2.9 fold increase relative to baseline creatinine & $\textless$ 0.5 \acrshort{mlkgh} $\geq$ 12 hours & \\
    \hline
    3 & 3 fold increase relative to baseline creatinine or absolute elevation to $\geq$ 4 \acrshort{mgdl} irrespective of baseline & 0.3 \acrshort{mlkgh} $\geq$ 24 hours or anuria $\geq$ 12 hours & initiation of dialysis \\
    \hline
\end{tabular}
 \label{tab:aki_staging}
 \caption{Staging of \acrlong{aki} according to \acrlong{kdigo} criteria} 
\end{table}

\subsubsection{Creatinine Baseline Definition} \label{sec:crea_baseline_definition}
There are several methods to define a baseline for \acrshort{scr}. A baseline is needed to compare \acrshort{scr} values against it at each point in time. Defining a baseline for \acrshort{scr} might vary due to data availability, data structure, local practices and standards. To take this into consideration, pyAKI contains a variety of different methods for defining a baseline of \acrshort{scr}: Using the minimum, mean and first value of either a fixed time frame at the start of the time series or a rolling time frame following the classification. Both time frames can have a self defined length. Also, a fixed value, e.g. a known preoperative \acrshort{scr} for surgical patients can be used. Lastly we also included a method to calculate a baseline creatinine value using a modification of the Cockcroft-Gault formula: 
\begin{equation}
\text{Cockcroft-Gault Creatinine Clearance, \acrshort{mlmin}} = \frac{(140 - \text{age}) \times \text{weight, \acrshort{kg}}  ( \times 0.85 \text{ if female})}{72 \times \text{\acrshort{scr}, \acrshort{mgdl}}}
\end{equation}
The Cockcroft-Gault formula is used to calculate the \acrfull{gfr} using the patients weight, gender and \acrshort{scr}~\cite{cockcroft_prediction_1976}.

The Cockcroft-Gault formula may be inaccurate depending on the patients weight. Therefore we further modified the formula to use the adjusted bodyweight of a patient if a height is provided for the patient~\cite{winter_impact_2012}. 

\begin{equation}
\text{\acrshort{scr}, \acrshort{mgdl}} = \frac{(140 - \text{age}) \times \text{weight, \acrshort{kg}}  ( \times 0.85 \text{ if female})}{72 \times \text{Cockcroft-Gault Creatinine Clearance, \acrshort{mlmin}}}
\end{equation}

Given this revised definition, it can be used to calculate an expected \acrshort{scr} under the assumption of a specified \acrshort{gfr}, which is 75\acrshort{mlmin} by default, but can be modified by the user.

\subsection{\acrshort{aki} Data Model}
\begin{figure}
    \centering
    \includegraphics[width=1\linewidth]{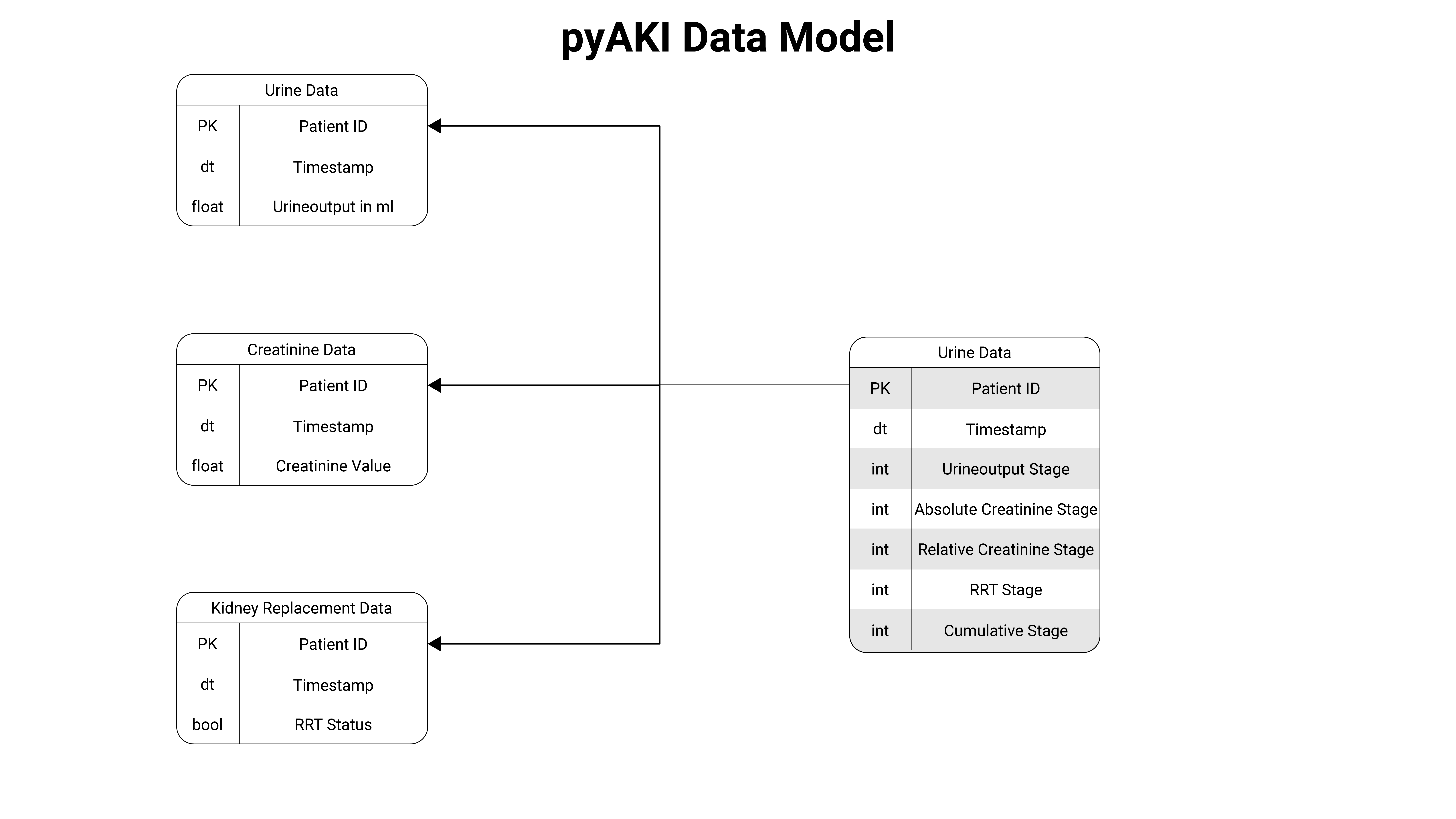}
    \caption{pyAKI Data Model. PK = Primary Key, dt = Datetime, bool = Boolean Value, int = Integer Value, float = Floating Point Value}
    \label{fig:pyAKI_ERD}
\end{figure}

For automated annotation of data, a standardized input format is required. We defined a minimal required set of variables and propose a data format consisting of three different data frames, each containing a subject identifier and timestamp for mapping to individual subjects. The required data frames are depicted in \autoref{fig:pyAKI_ERD}. We expect the data frames to contain hourly measured data of each individual subject, with no missing values.
At each subsequent point in time, \acrshort{kdigo} criteria were applied. \acrshort{aki} stages were evaluated separately, according to the pathways explained in \autoref{sec:aki_definition} and included in the output (see \autoref{fig:pyAKI_ERD}). The overall \acrshort{aki} stage is determined by the maximum stage in any of the three categories of the \acrshort{aki} definition. The output of the software contains each \acrshort{aki} defining criterion separately, together with the overall \acrshort{aki} stage at each hour of the patients \acrshort{icu} stay. As the input, the output also is formatted as a time series. \acrshort{aki} stages are evaluated at each point in time, in the case of our validation experiment, meaning at each hour of the \acrshort{icu} stay. Keeping these temporal relations within the data enables users to use it for further analysis in terms of \acrshort{aki} duration, recovery or temporal relation to other events.

\subsection{Software Development}\label{sec:software_development}
The pipeline was build using Python 3.11 (Python Software Foundation, Beaverton, Oregon). Starting from the data model for \acrshort{aki} mentioned above, we first developed interpolation methods for the data. Users should be able to parse real world data which is often sparse. By providing standardized imputation methods with high flexibility to fit the users requirements, we ensure an end-to-end pipeline from real life data to the desired output. These imputation techniques are used to up-sample and interpolate data to hourly intervals on which the \acrshort{aki} stages can be determined on. They are developed with a focus on customizability, where users can provide their own desired cutoffs for interpolation when data is missing over several hours. However, the imputations are not part of the core concept of pyAKI and are tested on synthetic data. Users have the flexibility to utilize pyAKI without employing imputation methods, allowing them to preprocess their data into the required hourly format according to their preferences. This ensures users have full control over their input data and acknowledges the heterogeneity of data sources.

\subsection{Validation Data Set}
To extract clinical data for our testing setup, we used the \acrshort{mimic}-IV demo version 2.2, which has been published in January 2023. This version of \acrshort{mimic} is an openly accessible subset of the full \acrshort{mimic} database which enabled us to share data publicly and provide a higher level of transparency in our process. The full \acrshort{mimic} dataset is not open for publication and requires special training and identification to be worked with. \acrshort{mimic} consists of more than 60,000 patients from the \acrshort{icu} of the Beth Israel Deaconess Hospital in Boston, United States of America. \acrshort{mimic}-demo provides a subset of 100 patients of \acrshort{mimic}-IV which are shared publicly over PhysioNet~\cite{johnson_alistair_mimic-iv_nodate}. 14 patients that had an \acrshort{aki} and one patient not fulfilling \acrshort{kdigo} criteria to serve as control, according to the definition implemented in the \acrshort{mimic} codebase, were sampled randomly from this data set and their \acrshort{scr}, \acrshort{uo} and dialysis-status was extracted from the appropriate tables.
The data was fitted according to our \acrshort{aki} data model as depicted in  \autoref{fig:pyAKI_ERD}. \acrshort{uo} was extracted in \acrshort{ml}, \acrshort{scr} in \acrshort{mgdl} or converted from \acrshort{mmol}/\acrshort{dl}. The need for dialysis was derived from the established \acrshort{mimic} concept~\cite{johnson_mimic_2018, noauthor_mit-lcpmimic-code_nodate} and converted to a Boolean variable indicating the dialysis status of the patient. The extracted data was up-sampled to hourly intervals and small windows ($< 6$ hours) of missing data were filled using forward filling in order to ensure continuous data for this experiment setup. For \acrshort{scr} baseline, we defined the baseline as a rolling window of the last seven days before each subsequent measurement for the relative \acrshort{scr} criterion, and a rolling window of the last 48 hours before each measurement for the absolute \acrshort{scr} criterion.

Physicians labels were provided by three physicians (Amini, Wida; Booke, Hendrik; Maidowski, Ludwig) working in the department for anaesthesiology and intensive care of the University Hospital Münster in Germany. They received a standardized initiation and instruction, which is appended in the supplements, consisting of an instruction on the \acrshort{kdigo} criteria and a presentation of unlabelled example data in the standardized format which is also used for pyAKI. The generated labels were then reviewed by an \acrshort{aki} expert panel consisting of two experienced physicians (Würdemann, Till; Weiss, Raphael) from the same department, who also received the standardized initiation and instruction for \acrshort{kdigo} criteria and are experienced in the care of patients with \acrshort{aki} and in \acrshort{aki} research. They reviewed the human annotations by their junior colleagues and annotations generated by pyAKI while the sources of the annotations (pyAKI or human) were anonymised and resolved conflicts between the two, resulting in a finally annotated data set. The full resulting data set, including labels by physicians and the result of the conflict resolution by the senior physicians are published alongside this paper ~\footnote{DOI: 10.17879/17988545762}. Not only did this method evaluate pyAKI against human performance, but it provided a standardized and labelled dataframe, which enables high quality testing in potential future development cycles. As mentioned above, \acrshort{aki} stages were evaluated at each hour of the \acrshort{icu} stay independently, keeping the temporal information within the data in order to enable users to leverage this for subsequent analysis. Therefore, each patient can be assigned with dozens and hundreds of \acrshort{aki} labels, depending on their length of observation.

\begin{figure}
    \centering
    \includegraphics[width=1\linewidth]{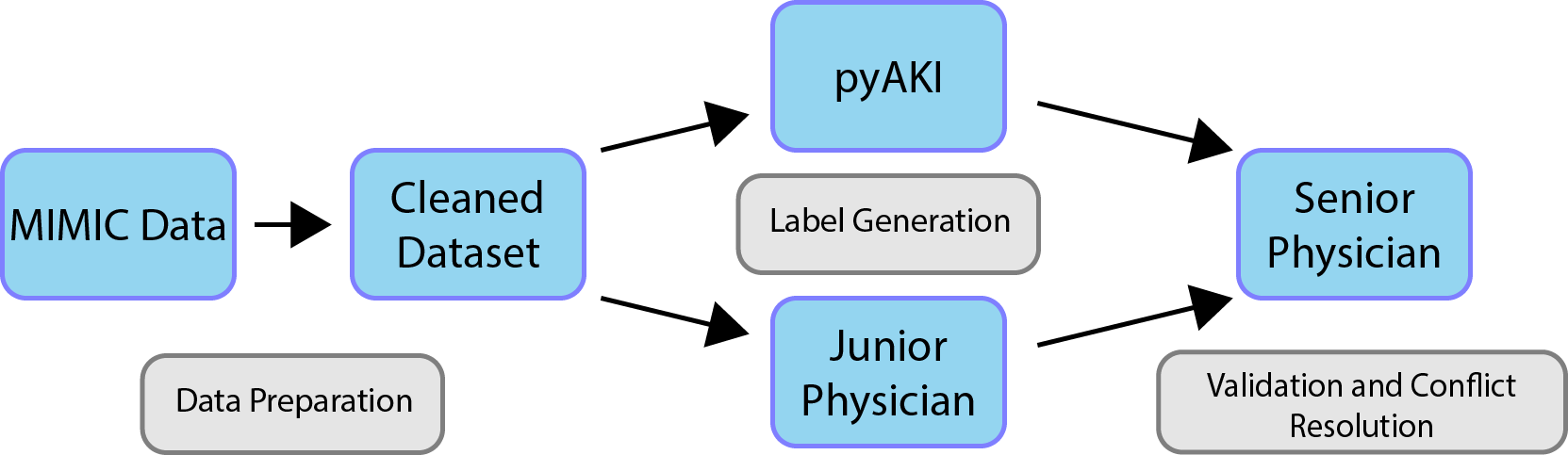}
    \caption{Workflow of validating the pyAKI pipeline.}
    \label{fig:pyaki_validation}
\end{figure}

\subsection{Pipeline Validation}\label{sec:pipeline_validation}
The entire workflow of the pipeline validation is depicted in Figure \ref{fig:pyaki_validation}. In order to validate the output of pyAKI, we employed a two step approach: First, every implemented probe, interpolation, class, method and function was tested thoroughly on artificial data. Tests are implemented in Python using the commonly applied unittest package, which is part of the standard Python library. All tests can be viewed and rerun locally by using our remote repository hosted on GitHub. Additionally, all tests are automatically executed when code changes happen to ensure correctness of the implemented algorithms. In the second step, the pipeline was validated using the novel validation data set. 

\section{Results}\label{sec:results}
PyAKI is a software package for the evaluation of time series data. Therefore, in the results, each point in time will be considered independently. In contrast to the standard expression in medical literature, we use "n" as measure for number of points in time, not as measure for number of patients.
\subsection{Patient Cohort}\label{sec:patient_cohort}
15 patients were randomly drawn from the \acrshort{mimic}-IV database. The entire data set, including all human labels, pipeline-generated labels and subsequent votings from the senior physicians can be found in the supplements. Overall, they included 2651 hours of patient data in the \acrshort{icu}. Mean length of ICU stay per patient was 176.73 hours. An overview of the descriptive parameters of the patients cohort is provided in table \autoref{tab:aki_and_accuracies}. All patients except for one fulfilled one or more criterion for \acrshort{aki} classification at least at one point in time of their \acrshort{icu} stay. The most common \acrshort{aki} stage per point in time was stage 0 (n=1623), followed by stage 3 (n=491). A complete overview of all \acrshort{aki}-stage counts per point in time can be found in table \autoref{tab:aki_and_accuracies}. The most common reason for the classification of a point in time as \acrshort{aki} was a reduction of urine output alone (n=508) and in combination with dialysis (n=179). Absolute elevations in \acrshort{scr} were also rather common alone and in combination with other factors (n=269), while relative elevations in \acrshort{scr} alone were less likely (n=28), yet more common in combination with other factors (n=66). 

\subsection{Accuracy}\label{sec:accuracy}
\begin{figure}
    \centering
    \resizebox{\textwidth}{!}{\includegraphics{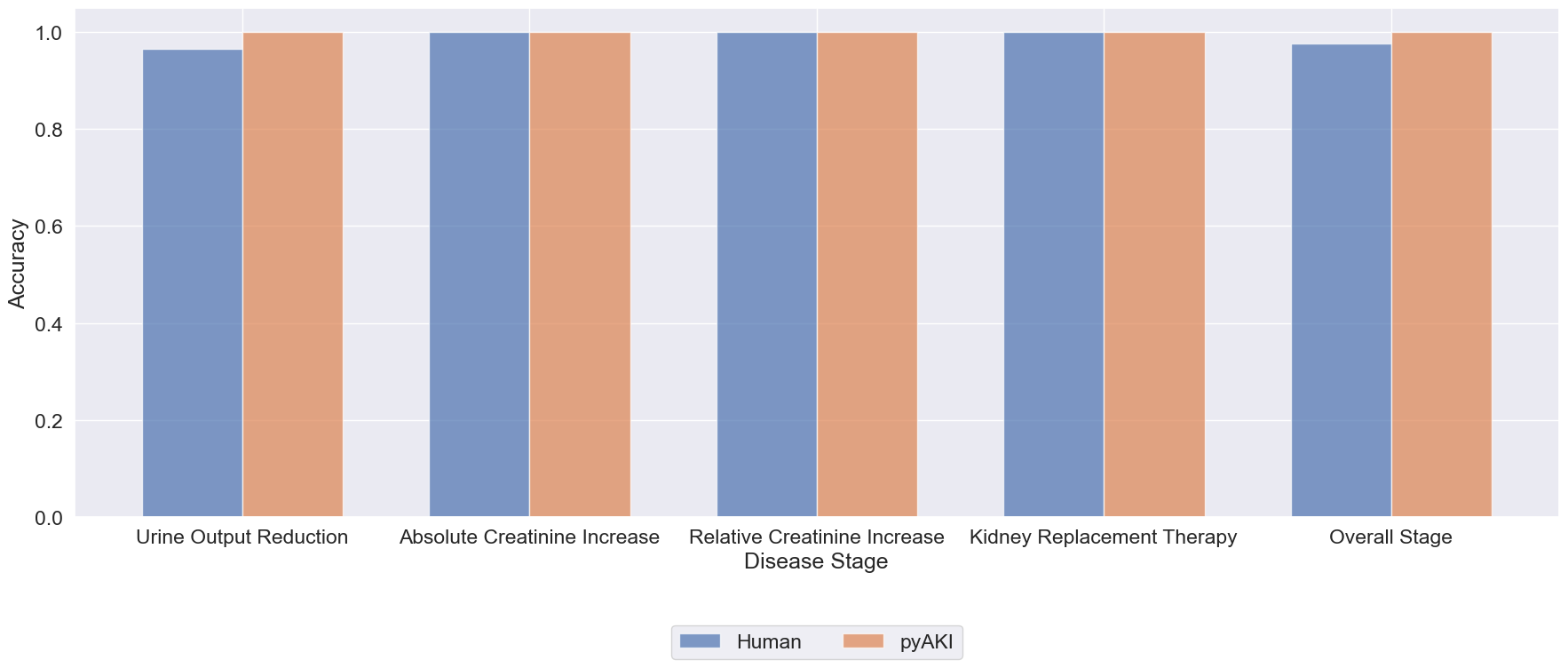}}
    \caption{Overall accuracy of human vs. pyAKI generated labels by classification method.}
    \label{fig:accuracies_by_method}
\end{figure}

An overview of all measured accuracies, categorized by each stage and each criterion of \acrshort{kdigo} classification, after comparing to the labels generated by the senior physicians is depicted in table \ref{tab:aki_and_accuracies}. Overall accuracy was high, both in the labels generated by physicians, as well as the labels generated by the pipeline. Accuracy across \acrshort{kdigo}-stages were also high (human vs. pyAKI): \acrshort{uo} was 0.9664 vs. 1.0 and relative \acrshort{scr} elevation was 0.9996 vs. 1.0. Accuracy for both dialysis-stage and absolute \acrshort{scr} elevation was 1.0 both in human labelling, as well as in labels generated by pyAKI. For overall \acrshort{aki}-stage, accuracy where 0.9771 for human labels and 1.0 for pyAKI generated labels. 
An overview of the accuracies across the different pathways of \acrshort{aki} staging is also provided in \ref{fig:accuracies_by_method}.

\subsection{\acrshort{aki} Characteristics}
Since time series data was used continuously for determination of \acrshort{aki} stages both by the experts, as well as the software pipeline, an evaluation of the accuracy in time of \acrshort{aki} diagnosis is possible. Timing of first \acrshort{aki} diagnoses were almost identical. Only in one case discrepancies occurred between the timing of \acrshort{aki} diagnoses, where the \acrshort{uo} stage was off by one hour by the physicians classification. Maximum \acrshort{aki} stages were evaluated with a high accuracy in all categories and the physicians labels and the labels determined by the algorithm matched in nearly all categories. Only a single case was determined incorrectly by the physicians. While the \acrshort{uo} stage was classified correctly as stage 1, the overall stage was determined as stage 0, due to a transmission error.

\begin{table}
    \begin{tabular}{ |p{3cm}||p{3cm}|p{3cm}|p{3cm}|p{3cm}|  }
\hline
 Category& \acrshort{aki} Label & No. of \acrshort{aki} per point in time & Human Accuracy&pyAKI Accuracy\\
 \hline
 \multirow{5}{4em}{\acrshort{uo}} &Overall&1726&0.9664&1.0\\
                                  &Stage 0&975&0.9991&1.0\\
                                  &Stage 1&108&0.9630&1.0\\
                                  &Stage 2&299&0.9030&1.0\\
                                  &Stage 3&344&0.9302&1.0\\
 \hline
 \multirow{3}{6em}{Absolute \acrshort{scr} Elevation}&Overall&2125&1.0&1.0\\
                                             &Stage 0&2106&1.0&1.0\\
                                             &Stage 1&223&1.0&1.0\\
                                             &Stage 3&109&1.0&1.0\\
 \hline
 \multirow{4}{6em}{Relative \acrshort{scr} Elevation}&Overall&2438&0.9996&1.0\\
                                                 &Stage 0&2311&0.9996&1.0\\
                                                 &Stage 1&65&1.0&1.0\\
                                                 &Stage 2&62&1.0&1.0\\
                                                 &Stage 3&0&1.0&1.0\\
 \hline
 \multirow{2}{4em}{Dialysis}&Overall&1914&1.0&1.0\\
                                &Stage 0&1610&1.0&1.0\\
                                 &Stage 3&304&1.0&1.0\\
                                 \hline 
 \multirow{4}{6em}{Overall \acrshort{aki} Stage}&Overall&2665&0.9771&1.0\\
                                                 &Stage 0&1665&0.9994&1.0\\
                                                 &Stage 1&203&0.9704&1.0\\
                                                 &Stage 2&306&0.9052&1.0\\
                                                 &Stage 3&491&0.9491&1.0\\
                                                 \hline
\end{tabular}
 \label{tab:aki_and_accuracies}
 \caption{Accuracy of human-assigned and pyAKI-generated labels across different acute kidney injury stages at each point in time. \acrshort{aki}=\acrlong{aki}, \acrshort{uo}=\acrlong{uo}, \acrshort{scr}=\acrlong{scr}}
\end{table}

\subsection{Misclassification}\label{sec:errors}
Errors in determining \acrshort{aki} stages can be grouped into three approximate categories: First, rounding errors, where errors due to rounding on digits after the decimal point occurred. Second, calculation errors, where simple miscalculation e.g. of the mean \acrshort{uo} over a given time window lead to a misclassification in the \acrshort{uo} stage. Third, misinterpretation or misreading of baseline values in the \acrshort{scr} category. Overall, most misclassifications by physicians were due to errors in calculation (n=57). Misinterpretation of baseline values caused one error and one error was due to an error in rounding values correctly.

\section{Discussion}\label{sec:discussion}
The proposed pyAKI software is the first standardized, validated and openly available tool for diagnosis of \acrshort{aki} in intensive care databases. Accordingly, the proposed data model is a first effort in harmonizing databases for subsequent analysis. Our pyAKI pipeline identified \acrshort{aki} patients with a high degree of agreement with ICU physicians. 
\newline
PyAKI outperformed the accuracy of a group of physicians on the intern level in most and equaled their performance in all metrics, including overall accuracy at each point in time, accuracy of determining the first \acrshort{aki} diagnosis, as well as accuracy in determining the highest \acrshort{aki} stage of a patient. However, these differences in accuracy were small and rarely had an effect on the overall classification of \acrshort{aki} stages. On one occasion, due to a transmission error, a patient was classified to not having \acrshort{aki} at all by the physicians classification, when in fact, pyAKI and the experts classification agreed, that the patient had \acrshort{aki}. Such minor mistakes, especially due to wrong transmission of single-category \acrshort{aki} determination to the overall \acrshort{aki} stage are typical for human level performance on complex time series data and should be avoidable by using a software pipeline.
\newline
Part of the superiority in performance of pyAKI opposed to the physicians might be due to the formation of the data. Tabular data is difficult to read, especially in long time series going over multiple hours or even days. This is especially true for urinary output data. Most currently used digital patient data management systems offer visualization frontends for users to facilitate easier managment for clinicians. However, our pipeline did not intend to outperform human level performance. As we have shown, human labels are of high quality and accurately represent \acrshort{aki} stages within time series data. By matching this quality, pyAKI is able to translate this quality to data sets on a large scale that would be resource intensive for human labelling.
\newline
Diagnosing \acrshort{aki} in large scale data sets is a time consuming tasks if it is done by human-generated labels and not performed algorithmically. Especially when doing so on both sparse and high frequency time series data. However, large sets of data are required where performance is dependent on the amount of available data, especially in machine learning. Past studies commonly relied on clinical labels or coded diagnoses for a patient within their hospital stay when investigating \acrshort{aki}. Prior studies demonstrated that such labelling is likely to create an under-representation of \acrshort{aki} in populations, especially for \acrshort{aki} stage 1~\cite{grams_performance_2014}. Khadzhynov et al. reported, that less than 30\% of \acrshort{aki} episodes were transferred to the medical documentation, depending on their stage. Wilson et al. also confirmed this finding in a large US population, where they even found a decreased 30 day mortality when \acrshort{aki} was not documented. This most likely occurred in less severe stages of illness. This relationship was reversed, when corrected for different scores of illness~\cite{wilson_impact_2013}. Even when working on the same data set, incidences of reported \acrshort{aki} can vary. In the EPI-AKI study, the authors were able to demonstrate that \acrshort{aki} across all stages occurs in more than 50\%, and moderate or severe \acrshort{aki} occurs in more than 30\% of all patients in the \acrshort{icu}~\cite{hoste_epidemiology_2015}. However, some studies, especially in the field of machine learning when working with large data sets and algorithmic implementation of \acrshort{kdigo} stages, reported incidences that differ substantially from this benchmark~\cite{rank_deep-learning-based_2020, schanz_under-recognition_2021, jiang_interpretable_2023, schmid_algorithm-based_2023}. This begs the question of why this deviation occurs and if it can all be attributed to local differences or if the algorithmic approach to classifying \acrshort{aki} might be flawed. Since most of the time implementations of such detection algorithms are not released with the publication, there is no way to reproduce the analysis and provide evidence for correctness of the output. There might be a lack of a standardized, well tested benchmark algorithm that can be referenced. By creating an expert-labelled validation data set and developing a toolbox upon it by testing it against that data, we aim to introduce this urgently needed standard.
\newline
To our knowledge, pyAKI is the first publicly available software pipeline for algorithmic determination of \acrshort{aki} stages on time series data. A recent study investigated the approach of algorithmic detection of \acrshort{aki} by applying \acrshort{kdigo} criteria~\cite{schmid_algorithm-based_2023}. In this study, the authors included 21045 cases of post-cardiac surgery patients from 2012 to 2022 and applied \acrshort{kdigo} criteria in an automated, algorithmic fashion using the programming language R (R Foundation for Statistical Computing, Vienna, Austria)~\cite{r_core_team_r_2022}. However, the authors did not evaluate their calculations against human level performance or another standardized method of classification. While detecting \acrshort{aki} rates of over 60\% in postoperative \acrshort{aki}, which is close to the expected \acrshort{aki} rates described by Hoste et al. and Zarbock et al.~\cite{hoste_epidemiology_2015, zarbock_epidemiology_2023}, the authors to our knowledge and to this date, did not publicise their algorithm for \acrshort{aki} detection which prohibits validation of their findings.
\newline
The \acrshort{kdigo} guidelines do not provide a clear recommendation on what qualifies as baseline \acrshort{scr} for defining \acrshort{aki} stages based on \acrshort{scr} and several methods are used by clinicians. To account for this, we have incorporated the most commonly used methods as mentioned under section \ref{sec:crea_baseline_definition}: Users can choose the minimum, mean or first value of a time window at the start of the time series, as well as a rolling window following the classification. The length of the window can be self defined in both cases. Users can also choose to provide a fixed value if a known baseline is used (e.g. preoperative \acrshort{scr}), as well as choosing a method of calculating the baseline under the assumption of a user defined \acrshort{gfr} via the Cockcroft-Gault formula for use in pyAKI and allow the user to choose the most appropriate method based on individual preference. 
By providing a set of different formulas and methods for baseline calculation  we intend to increase transparency across study applications. The definition of a baseline might vary due to a number of reasons. Most commonly, especially in surgical patients, a preoperative or pre-hospitalisation SCr might be the most appropriate approximation of a baseline creatinine value. This will often not be accessible though, especially when working only with \acrshort{icu} data. Our pipeline offers the user the possibility to be fully transparent about the exact implementation of a baseline SCr within their data. Using the Cockcroft-Gault formula as a possibility of baseline definition might be considered the most unsuitable method, as the \acrshort{gfr} can be inconsistent and biased, especially in \acrshort{icu} patients, patients affected by chronic kidney disease or patients using diuretics. Therefore this possibility should be considered a last option if no other way of baseline definition is available. It might be a valid option if a study population is sure to be without patients affected by chronic kidney disease and no other method of baseline definition seems applicable. 
\newline
Our pipeline is ready for use, however it has some limitations. First, due to limitations in the programming language Python and other programming languages as well, floating point errors might lead to a misclassification of \acrshort{aki} both in the \acrshort{scr} as well as the \acrshort{uo} track. These should only affect a very limited amount of edge cases with questionable clinical relevance. Another limitation is that according to \acrshort{kdigo} criteria, in patients under 18 years of age, a reduction in \acrshort{gfr} below 35 \acrshort{mlmin} per 1.73 m\textsuperscript{2} is considered an \acrshort{aki} stage 3. Since we currently did not have openly accessible and publishable real world data from individuals younger than 18 years, we did not validate the pipeline on this criterion and exclusively validated it on patients over the age of 18. Finally, heterogeneity in electronic health record databases may require modifications of the data structure before pyAKI can be implemented in some cases. These modifications might introduce bias to the data analysis that are beyond the capabilities of pyAKI. As in most software toolboxes, the quality of the output is still dependent on the quality of the input. 
\newline
In conclusion, pyAKI provides a standardized, validated and openly accessible tool for classifying time series data with \acrshort{kdigo} \acrshort{aki} labels. These labels achieve and even outperform human level accuracy and can provide a new standard of the definition for \acrshort{aki} in large scale data sets.
\newpage
\section{Acknowledgements}
Christian Porschen is supported by the Deutsche Forschungsgemeinschaft (DFG, German Research Foundation) – 493624047 (Clinician Scientist CareerS Münster).
\newline
\newline
Thilo von Groote is supported by the Deutsche Forschungsgemeinschaft (DFG, German Research Foundation) – 493624047.
\section{Conflicts of Interest}
None.
\newpage
\printglossaries

\newpage
\printbibliography

\end{document}